# Recursive State Inference for Linear PASFA

**Vishal Rishi[1]**
[1]Fast Code AI
Bangalore, India
vishalr@fastcode.ai

***Abstract*** - Slow feature analysis (SFA), as a method for learning slowly varying features in classification and signal analysis, has attracted increasing attention in recent years. Recent probabilistic extensions to SFA learn effective representations for classification tasks. Notably, the Probabilistic Adaptive Slow Feature Analysis models the slow features as states in an ARMA process and estimate the model from the observations. However, there is a need to develop efficient methods to infer the states (slow features) from the observations and the model. In this paper, a recursive extension to the linear PASFA has been proposed. The proposed algorithm performs MMSE estimation of states evolving according to an ARMA process, given the observations and the model. Although current methods tackle this problem using Kalman filters after transforming the ARMA process into a state space model, the original states (or slow features) that form useful representations cannot be easily recovered. The proposed technique is evaluated on a synthetic dataset to demonstrate its correctness.

## 1. Introduction

Slow Feature Analysis (SFA) [3] is a dimensionality reduction and feature extraction technique designed to identify slowly varying features from rapidly changing input signals. The fundamental principle behind SFA is that slowly changing aspects of data often contain meaningful information about the underlying processes. Since its introduction, SFA has found applications in various domains including computer vision, robotics, and signal processing. Traditional SFA approaches focus on finding projections that minimize the temporal variation of features while ensuring decorrelation and unit variance constraints. While these approaches have demonstrated success, they typically lack probabilistic interpretations, making them sensitive to noise and outliers. To address these limitations, probabilistic extensions to SFA have been developed in recent years. Among these extensions, Probabilistic Adaptive Slow Feature Analysis (PASFA) [1] provides a robust framework by modeling slow features as states in an Autoregressive Moving Average (ARMA) process. PASFA effectively estimates the model parameters from observations, offering advantages in terms of adaptability and noise handling. However, a significant challenge in PASFA is the efficient inference of states (slow features) from observations once the model has been learned. The standard state inference method in PASFA uses the estimated posterior distribution over the states. But the posterior does not consider the state dynamics and hence, the state estimates are not optimal. In this paper, we propose a recursive extension to linear PASFA that directly addresses this challenge. Our approach enables recursive MMSE estimation of states evolving according to an ARMA process, given both the observations and the model parameters. One of the approaches for recursive state inference typically transforms the ARMA process into a state-space model and applies Kalman filtering techniques [2]. While this transformation facilitates the filtering process, it introduces several issues: 1) The transformation increases the dimensionality of the state vector, leading to higher computational costs. 2) The resulting state estimates do not directly correspond to the original slow features, complicating interpretation. 3) The recovery of original slow features from the augmented state estimates introduces additional error. By avoiding the transformation to state-space models, our method preserves the interpretability and utility of the original slow features.

## 2. Related Work

Slow Feature Analysis (SFA) [3], originally introduced by Wiskott and Sejnowski, is an unsupervised learning method inspired by the temporal stability of visual processing, aiming to extract features with minimal temporal variation under decorrelation and unit variance constraints. Extensions such as graph-based SFA and nonlinear variants using kernels or neural networks have broadened their applicability to complex temporal and nonlinear data [5, 6]. Probabilistic formulations marked a significant advancement, with Turner and Sahani [4] framing SFA as inference in a linear dynamical system, leading to Probabilistic SFA (PSFA), which treats slow features as latent variables with temporal priors. Probabilistic Adaptive SFA (PASFA) [1] further generalizes this by modeling slow features as ARMA processes, enabling more flexible temporal dynamics and using maximum likelihood via EM algorithms for learning. However, inferring states in ARMA models remains challenging. While traditional methods transform ARMA models into state-space form to apply Kalman filtering [2], they often face efficiency or scalability issues, especially in high-dimensional SFA settings. Also, the recovery of original latent states from the augmented state estimates introduces additional error.

## 3. Methodology

### 3.1. Linear PASFA Framework

In the linear PASFA framework, features are assumed to be dynamically driven by inherent (endogenous) uncertainties. For $j \in \{1, ..., P\}$, the $j^{th}$ latent signal $x_j[k]$ follows,

$$x_j[k] = H_j(q^{-1})e_j[k] \quad (1)$$

Since we assume an ARMA process, $H_j(q^{-1}) = \dfrac{1 + \sum_{i=1}^{M} b_{ji} q^{-i}}{1 - \sum_{i=1}^{L} a_{ji} q^{-i}}$

We have assumed that all signals are zero at negative time instants. $e_j[k] \sim N(0, \sigma_j^2)$ is the process noise and $q^{-1}$ is the backward shift operator. Equation (1) can be written in the difference equation form as follows,

$$x[k] = \sum_{l=1}^{L} A_l x[k-l] + \sum_{l=1}^{M} B_l e[k-l] + e[k] \quad (2)$$

$$A_l = diag(a_{1l}, a_{2l}, ..., a_{Pl}), \quad \forall l \in \{1, ..., L\} \quad (3)$$

$$B_l = diag(b_{1l}, b_{2l}, ..., b_{Pl}), \quad \forall l \in \{1, ..., M\} \quad (4)$$

$$\Sigma_e = diag(\sigma_1^2, ..., \sigma_P^2) \quad (5)$$

Here, $x[k] = [x_1[k], ..., x_P[k]]^T$ and $e[k] = [e_1[k], ..., e_P[k]]^T$. The observation signal $y[k] \in R^{M \times 1}$ follows the linear generative equation,

$$y[k] = Cx[k] + \varepsilon[k] \quad (6)$$

$\varepsilon[k] \sim N(0, \sigma_\varepsilon^2 I_{Q \times Q})$ is the measurement noise. Given a series of observation signals $y[k]_{k=0:T-1}$, the parameters $A_{l=1:L}$, $B_{l=1:M}$, $\Sigma_e$, $C$, and $\sigma_\varepsilon^2$ can be estimated using a variational EM algorithm [1]. The estimated parameters ensure the second-order stationarity of $x[k]$.

### 3.2. Proposed Algorithm

In this section, we derive the algorithm for recursive MMSE estimation of states evolving according to equation (2), given the observation signals $y[k]_{k=0:T-1}$, and the parameters $A_{l=1:L}$, $B_{l=1:M}$, $\Sigma_e$, $C$, and $\sigma_\varepsilon^2$. We introduce two definitions,

$$\hat{x}[k|k-l] = E(x[k] | y[0], ..., y[k-l]), \quad l = 0, ..., N \tag{7}$$

$$P[k, k-l|k-m] = E\left[\left(x[k] - \hat{x}[k|k-m]\right)\left(x[k-l] - \hat{x}[k-l|k-m]\right)^T\right] \tag{8}$$

$$l = 0, ..., m, \text{ and } m = 0, ..., N$$

Here, $N = (L, M+1)$, $\hat{x}[k|k-l]$ is the MMSE estimator of $x[k]$ given the observation signals till the $(k-l)^{th}$ time instant and $P[k, k|k-l]$ ($P[k|k-l]$ in shorthand notation) is the covariance matrix of the estimation error $\left(x[k] - \hat{x}[k|k-l]\right)$. From equation (2),

$$\hat{x}[k|k-N] = \sum_{l=1}^{L} A_l \hat{x}[k-l|k-N] \tag{9}$$

$$\hat{x}[k|k-l] = \hat{x}[k|k-l-1] + F[k, k-l]\hat{e}[k-l|k-l-1], \quad l = 0, ..., N-1 \tag{10}$$

$F[k, k-l]$ is the filter gain and $\hat{e}[k-l|k-l-1] = y[k-l] - C\hat{x}[k-l|k-l-1]$. Equation (10) is a generalization to the recursive update equation derived in the Kalman filter. From equations (9) and (10), recursive relations for the error covariance matrix are derived.

$$P[k|k-N] = \sum_{i=1}^{L} A_i P[k-i|k-N] A_i^T + \Sigma_e \left(I + \sum_{i=1}^{M} B_i B_i^T\right) + 2 \sum_{i,j}^{L,M} A_i \gamma_{ex}[i-j] B_j^T \tag{11}$$

$$P[k, k-l|k-N] = \sum_{i=1}^{L} A_i P[k-i, k-l|k-N] + \sum_{i=1}^{M} B_i \gamma_{ex}[l-i], \quad l = 1, ..., N \tag{12}$$

$\gamma_{ex}[l] = E\left(e[k]x[k-l]^T\right)$ is the cross-covariance function between $e[k]$ and $x[k]$.
For $l = 0, ..., m, \text{ and } m = 0, ..., N-1$,

$$P[k, k-l|k-m] = P[k, k-l|k-m-1] - P[k, k-m|k-m-1]C^T F^T[k-l, k-m] - F[ \tag{13}$$

The optimal filter gains are obtained by solving the following MMSE optimization objective,

$$F^*[k, k-l] = P[k|k-l], \quad l = 0, ..., N-1 \tag{14}$$

Substituting equation (13) in (14) and solving the optimization objective for $F[k, k-l]$,

$$F^*[k, k-l] = P[k, k-l|k-l-1]C^T \left[CP[k-l|k-l-1]C^T + \sigma_\varepsilon^2 I\right]^{-1}, \quad l = 0, ..., N-1 \tag{15}$$

Thus, $\hat{x}[k|k]_{k=0:T-1}$ form the MMSE estimates of $x[k]_{k=0:T-1}$, given the observations and the data-generating model.

## 4. Experiments and Results

To evaluate the effectiveness of our proposed recursive extension to PASFA, we generated a synthetic dataset that follows the data-generating process in equations (2) and (6). The train and test dataset consists of 2000 (= $T$) time points each, with the dimensionality of the observations ($P$) and the slow features ($Q$) to be 1. The synthetic data was generated according to the following model:

$$x[k] = 0.0935x[k-l] + e[k] + 0.2416e[k-l] \qquad (16)$$

$$y[k] = 0.5253x[k] + \varepsilon[k] \qquad (17)$$

Here, $e[k] \sim N(0, 0.7577)$ and $\varepsilon[k] \sim N(0, 0.2955)$. All the parameters were sampled randomly to avoid selection bias. We evaluated our proposed algorithm against the standard inference implementation of PASFA. Given the train observation signals, we estimate all the model parameters using the method mentioned in [1]. Once the estimated model is obtained, state inference is performed using the test observation signals. The performance metrics include the Mean Squared Error (MSE) and the correlation coefficient between the true and estimated slow features. Our experiments demonstrate that the proposed recursive algorithm effectively tracks the true slow features, capturing both their magnitudes and temporal patterns (refer to Figure 1). Quantitative 10-fold cross-validation shows that our method achieves significantly better MSE and correlation coefficient than the standard inference approach for PASFA (refer to Table 1).

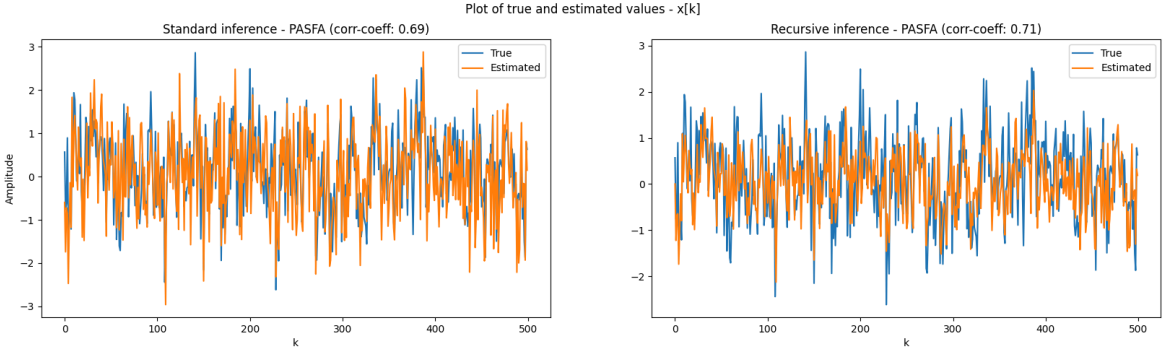

Fig. 1: Plot of true and estimated slow features (correlation coefficient in brackets)

Table 1: Evaluation Metrics – over 10-fold cross-validation (standard deviation in brackets)

| Method | MSE | Correlation Coefficient |
|---|---|---|
| Standard Inference - PASFA | 0.6005 ($\pm$0.018) | 0.696 ($\pm$0.017) |
| Recursive Inference - PASFA | 0.4871 ($\pm$0.011) | 0.712 ($\pm$0.019) |

## 5. Conclusions

This paper introduces a recursive extension to linear Probabilistic Adaptive Slow Feature Analysis (PASFA) for efficient state inference, leveraging the ARMA formulation of slow feature dynamics to avoid complex state-space transformations and preserve the interpretability of original slow features. The core contribution is a recursive MMSE state estimation algorithm for ARMA processes within the linear PASFA framework, which demonstrates improved accuracy on synthetic data compared to standard PASFA inference. However, the method assumes linear observation models and fixed ARMA parameters, and its evaluation is currently limited to synthetic datasets. Future work will address these limitations by incorporating nonlinear observation models via advanced filtering techniques, enabling joint parameter and state estimation for online adaptation, validating performance on real-world datasets, and exploring integration with deep learning architectures.